\documentclass[utf8]{frontiersSCNS} 

\usepackage{url,hyperref,lineno,microtype,subcaption}
\usepackage[onehalfspacing]{setspace}
\usepackage[titlenumbered,ruled,linesnumbered]{algorithm2e}
\usepackage{algpseudocode}

\def\firstAuthorLast{Wang {et~al.}} 
\def\Authors{Siyu Wang\,$^{1,*,\dagger}$, Yuanjiang Cao\,$^{1,\dagger}$, Xiaocong Chen\,$^{1}$, Lina Yao\,$^{1}$, Xianzhi Wang\,$^{2}$, and Quan Z. Sheng\,$^{3}$}
\def\Address{$^{\dagger}$These authors contribute equally to this work and share the first authorship

$^{1}$School of Computer Science and Engineering, University of New South Wales, Sydney, NSW, Australia 

$^{2}$School of Computer Science, University of Technology Sydney, Sydney, NSW,  Australia 

$^{3}$Department of Computing, Macquarie University, Sydney, NSW,  Australia }

\def\corrAuthor{Lina Yao}
\def\corrEmail{lina.yao@unsw.edu.au}

\begin{document}
\onecolumn
\firstpage{1}

\title[Adversarial Robustness of DRL RS]{Adversarial Robustness of Deep Reinforcement Learning based Dynamic Recommender Systems} 

\author[\firstAuthorLast ]{\Authors} 
\address{} 
\correspondance{} 
\extraAuth{}

\maketitle

\begin{abstract}
Adversarial attacks, e.g., adversarial perturbations of the input and adversarial samples, pose significant challenges to machine learning and deep learning techniques, including interactive recommendation systems.
The latent embedding space of those techniques makes adversarial attacks difficult to detect at an early stage. Recent advance in causality shows that counterfactual can also be considered one of ways to generate the adversarial samples drawn from different distribution as the training samples.
We propose to explore adversarial examples and attack agnostic detection on reinforcement learning-based interactive recommendation systems. We first craft different types of adversarial examples by adding perturbations to the input and intervening on the casual factors. Then, we augment recommendation systems by detecting potential attacks with a deep learning-based classifier based on the crafted data. Finally, we study the attack strength and frequency of adversarial examples and evaluate our model on standard datasets with multiple crafting methods. Our extensive experiments show that most adversarial attacks are effective, and both attack strength and attack frequency impact the attack performance. The strategically-timed attack achieves comparative attack performance with only 1/3 to 1/2 attack frequency. Besides, our black-box detector trained with one crafting method has the generalization ability over several other crafting methods.
\end{abstract}

\section{Introduction}
Recommendation systems are an effective means of alleviating information overload for Internet users.
They generally filter out those less irrelevant ones from massive items of choice to improve user experience in multiple scenarios.
Traditional recommendation systems extract features about user preferences, items, and users' past interactions with items to conduct content-based, collaborative, or hybrid recommendation~\citep{adomavicius2005toward, zhang2019deep}. These models have not considered the changes in user preferences over time.
In this regard, interactive recommendation systems emerge to capture personalized user preference dynamics.
Generally, interactive recommendation systems cater to users' dynamic and personalized requirements by improving the rigid strategy of conversational recommendation systems~\citep{mahmood2007learning,thompson2004personalized, Taghipour2008hybrid}.
In recent years, they have been attracting increasing attention and employed in leading companies (e.g., Amazon, Netflix, and YouTube) for personalized recommendations.

Interactive recommendation systems can be considered a decision-making process where the system chooses an optimal action in each discrete step to maximize the user response evaluation. Common practices to model the interactions between recommendation systems and users include Multi-Armed Bandit (MAB) or Reinforcement Learning (RL).  
The former views the action choice as a repeated single process while the latter considers immediate and future rewards to better model long-term user preference behaviors. In RL-based recommender systems, a Markov Decision Process (MDP) agent estimates the value function by both action and state rather than merely by action as done by MAB.

However, small disturbances in the input data may fool the above practices~\citep{Szegedy2013Intriguing, Goodfellow2014Explaininga}.
Small imperceptible noises, such as adversarial examples, may increase prediction error or reduce reward in supervised and RL tasks --- the input noise can be transferred to attack different parameters even different models, including recurrent network and RL~\citep{gao2018black,Huang2017Adversarial}.
Besides, the vector representations of entity/relation embedding of the input of RL-based recommendation models make it challenging for humans to tell the true value or dig out the real issues in the models.

Recently,~\cite{browne2020semantics} points out that counterfactual reasoning can be used to generate the adversarial samples. From the perspective of causal inference, counterfactual data can be generated by intervening in some aspect of input data. Both perturbations and counterfactual reasoning target the state space by introducing noise. Attackers can easily leverage such characteristics of embedding vectors to disrupt recommendation systems silently. Therefore, it is important to study attack and defense methods for RL-based recommendation systems.

This work aims to develop a general detection model to detect attacks and increase the defense ability and robustness, which provides a practical strategy to overcome the dynamic 'arm-race' of attack and defense in the long run. The problem is nontrivial due to three reasons.
First, online attacks are inherently difficult to track or predict.
Second, man-in-middle methods can attack the interactions between recommendation systems and users in web applications, giving opportunities for malicious people to disrupt recommendation systems in either a white-box or a black-box way. Third, the huge number of actions in RL-based recommendation systems poses a barrier to detecting user feedback since the exhaustively numerous items and users embedding vectors are not feasible to find the abnormal inputs. We propose an attack-agnostic detection model against adversarial examples for RL-based recommendation systems to overcome the above challenges.
To the best of our knowledge, this is the first work that focuses on the adversarial detection of RL-based Recommendation Systems. We make the following contributions:
\begin{itemize}
    \item We systematically investigate different types of adversarial attacks and detection approaches focusing on reinforcement learning-based recommendation systems and demonstrate the effectiveness of the designed adversarial examples and strategically-timed attack. 
   \item We propose an encoder-classification detection model for attack-agnostic detection. The encoder captures the temporal relationship among sequence actions in reinforcement learning. We further use an attention-based classifier to highlight the critical time steps out of a large interactive space.
   \item We empirically show that even small perturbations or counterfactual states can significantly reduce most attack methods' performance.
   Our statistical validation shows that multiple attack methods generate similar actions of the attacked system, providing insights for improving the efficacy of the detection performance. 
\end{itemize}

\section{Related Work}
\vspace{1mm}\noindent\textbf{RL-based interactive recommendation.}
Reinforcement learning is a popular approach to interactive recommendation.
Traditional research applies Q-learning~\citep{Taghipour2007Usage-Based, Taghipour2008hybrid} and Markov Decision Process~\citep{Mahmood2009Improving} to web recommendation and conversational recommendation problems.
\cite{mahmood2007learning} first introduce reinforcement learning into interactive recommendation by modifying MDP.
Since then, deep learning has inspired more interest in interactive recommendation.
For example,~\cite{christakopoulou2018q} employ reinforcement learning to improve feedback quality in interactive recommendation;~\cite{chen2019large} adopt policy gradient to improve the scalability of interactive recommendation.

\vspace{1mm}\noindent\textbf{Adversarial attacks.}
\cite{Szegedy2013Intriguing} first find that hardly perceptible perturbation can cause erroneous outputs of a convolutional neural network on image classification tasks.
\cite{Goodfellow2014Explaininga} exploit this topic further and incorporate the Fast Gradient Sign Method to attack neural networks; they find the strong linear component of neural networks have a relationship with adversarial attack success.
Further studies involve selecting attack points \citep{papernot2016limitations}, untargeted \citep{moosavi2016deepfool} and targeted attack models\citep{Lin2017Tacticsc}, and experimental comparisons of random noise and adversarial examples \citep{Kos2017Delving}.
Specifically,~\cite{Lin2017Tacticsc} design strategically-timed attacks and craft deceptive images to induce the agent to make the desired actions.
\cite{browne2020semantics} argue that counterfactual explanations produce adversarial examples in DL research, which modify the input to cause misclassification of the network.
\cite{Huang2017Adversarial} explore the adversarial attack deep Q network in video game playing and conclude that retraining with adversarial examples can make the network more robust.
Another thread of research models adversarial attacks into environments for robust adversarial training. They either regard the attack as a destabilizing force to break the balance of agents in 3D scenarios~\citep{Pinto2017Robusta} or develop adversarial agents in multi-agent tasks during reinforcement learning~\citep{Gleave2019Adversariala}. 
Generally, creating adversarial examples helps reduce the reward of on DQN and DDPG \citep{Pattanaik2017Robust}, and a detection method can help better explore the potential of adversarial examples and make agents more robust in a dynamic process.

\vspace{1mm}\noindent\textbf{Adversarial example detection.}
Many adversarial detection methods are vulnerable to loss functions targeted to fool them~\citep{Carlini2017Adversariala}.
Bendale et al. \cite{Bendale2016Open} present OpenMax to estimate the probability of data input from unknown classes explicitly.
Since then, researchers have proposed statistical approach~\citep{Hendrycks2016Early}, binary classification approach~\citep{Metzen2017Detecting}, outlier detection approach~\citep{Grosse2017(Statistical)}, and history queries-based approach~\citep{Chen2019Stateful} to detect adversarial examples.
Our work differs from~\cite{Chen2019Stateful} in exploiting the nature of reinforcement learning besides query-based black-box attacks.
\begin{figure*}[!ht]
  \centering
    \includegraphics[width=1.0\textwidth]{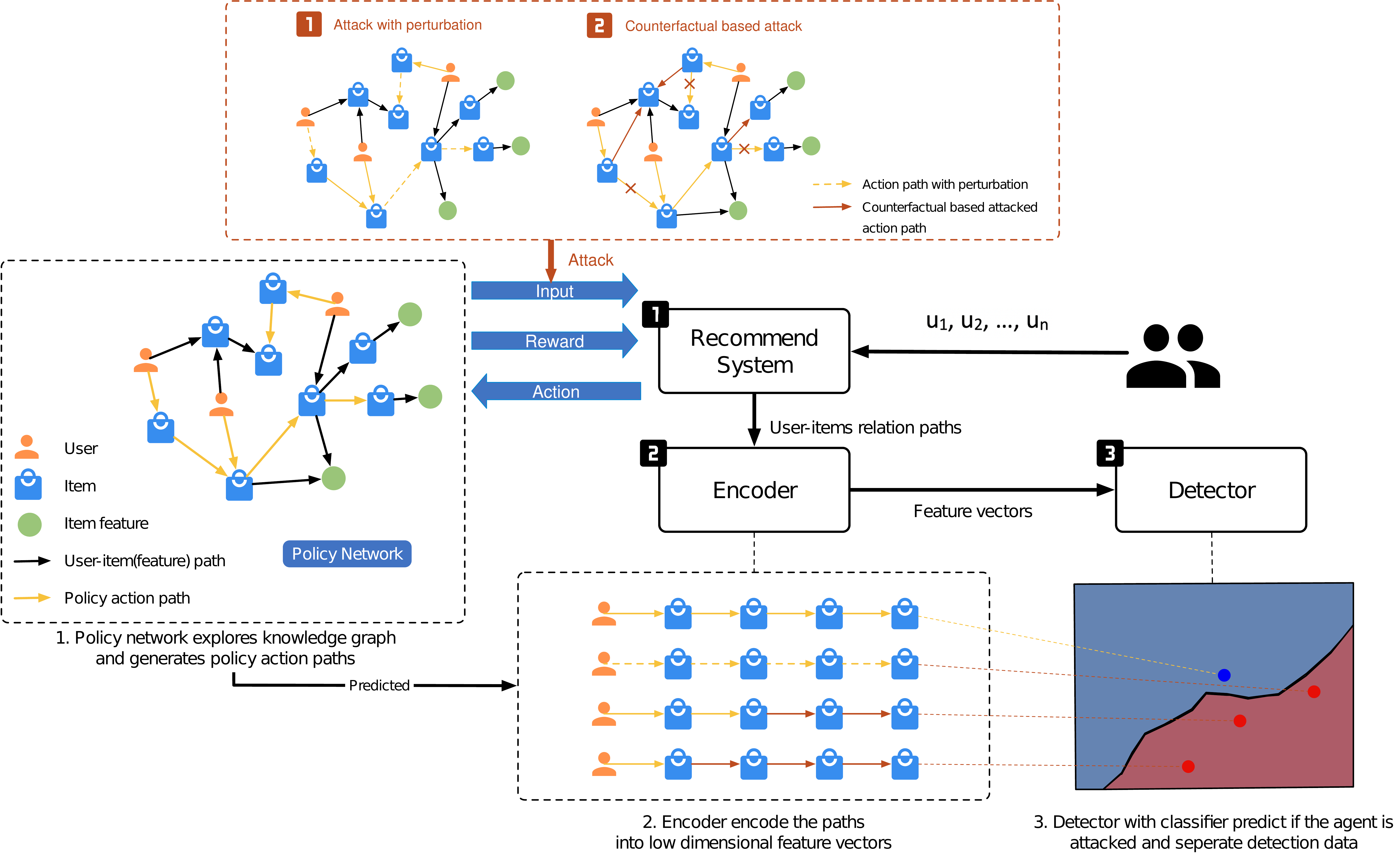}
  \centering
  \caption{
  Our proposed Adversarial Attack and Detection Approach for RL-based Recommender Systems.
  }
  \label{fig-attention-0}
\end{figure*}

\section{Methodology}
\label{method}
This section introduces the components of an RL-based recommendation system, attack techniques that generate adversarial examples, and our scheme to detect black-box adversarial attacks.

\subsection{RL-based Interactive Recommendation}
\label{problem-definition}
Interactive recommendation systems suggest items to users and receives feedback. Given a user $u_{j} \in U= \{u_0, u_1, u_2, ..., u_n\}$, a set of items $I = \{i_0, i_1, i_2, ..., i_n\}$, and the user feedback history $i_{k_1}, i_{k_2}, ..., i_{k_{t-1}}$, the recommendation system suggests a new item $i_{k_{t}}$.
This problem can be regarded as a Markov Decision Process, which comprises the following:
\begin{itemize}
    \item State ($s_t$): a historical interaction between user and the recommendation system computed by an embedding or encoder module.
    \item Action ($a_t$): an item or a set of items recommended by the RL agent.
    \item Reward ($r_t$): a variable related to user's feedback to guide the reinforcement model towards true user preference. 
    \item Policy ($\pi(a_t | s_t)$): a probabilistic model consisting of an estimation and action generation parts. The training process aims to obtain an optimal policy for recommendation.
    \item Value function ($Q(s_t, a_t)$): the agent's prediction of reward of current recommended item $a_t$. 
\end{itemize}
The reinforcement agent could be an Actor-Critic Algorithm that consists of a critic network and an actor network~\citep{xian2019reinforcement}. The attack model may generate adversarial examples using either the critic network~\citep{Huang2017Adversarial} or the actor network~\citep{Pattanaik2017Robust}.

\subsection{Attack Model}
\label{method-attack}
\vspace{1mm}\noindent\textbf{FGSM-based attack.} We define an adversarial example as a little perturbation $\delta$ added onto the benign examples $x$ to reduce the cumulative reward of a reinforcement learning system.
Suppose $x$ is a sequence of feature vectors piped into reinforcement learning model $\pi(s_t)$; $x$ can be a composition of embedding vectors of users, relations, and items~\citep{xian2019reinforcement}, or a feature vector encoded user and item information~\citep{chen2019large}.
Unlike perturbations on images or texts, $\delta$ can be large in interactive recommendation systems due to the enormous manual work to check the embedding vectors or feature vectors of massive users and items. We define an adversarial example as follows:
\begin{equation}
\begin{aligned}
    & \min\limits_{\delta} R_T = \sum\limits_{t=1}\limits^{T} r_t\\
    & r_t = Q(s_t + \delta,\ a_t)\\
    & a_t = \pi^{*}(a_t | s_t + \delta)\\
    & \text{subject to}\quad S(s_t, s_t+\delta) \leq l
\end{aligned}
\end{equation}
where $R_T$ is the total reward of the recommendation agent, $T$ is the length of a time step, $\pi^{*}$ is the optimal policy learned by the training process, $S$ ($<l$) is a similarity metric that measures the distance between benign and adversarial examples.
$S$ is commonly defined as $l_p$ bounded perturbation, or $|\delta|_p$~\citep{carlini2019Onevaluating}. The computation of $\delta$ determines the method of attack. We aim to build a model with the generalization ability to detect examples from unknown adversarial distributions. Thus, we adopt three attack methods to validate the detection model performance: FGSM~\citep{Goodfellow2014Explaininga} and its variant~\citep{Huang2017Adversarial}, JSMA~\citep{papernot2016limitations}, and Deepfool ~\citep{moosavi2016deepfool}. FGSM can be presented as follows:

\begin{equation}
\label{eqt-l-inf}
\begin{aligned}
\delta_{inf} = \epsilon\ sign(\nabla_{s_t}J(Q_t, Q_{t-1} + r_t))
\end{aligned}
\end{equation}

where $J$ is the loss function, $Q_t$ is the critic function $Q(s_t, a_t)$. Optimizing $J$ will lead to the critic value $Q$ satisfying the Bellman equation. The FGSM method uses the gradient of the loss function, which can be computed efficiently, thus requiring a small amount of additional computation.

To construct a detection model with the generalization ability, we train the detection model with FGSM examples and conduct the detection using other perturbation methods. We adopt the two norm variations in~\citep{Huang2017Adversarial} and define the norm constraint of perturbations as follows:
\begin{equation}
\label{eqt-l1}
\begin{aligned}
    & \delta_{2} = \epsilon\sqrt{d}*\frac{\nabla_{s_t}J(Q_t, Q_{t-1} + r_t)}{||\nabla_{s_t}J(Q_t, Q_{t-1} + r_t)||_2}\\
    & \delta_{1} = \text{perturb highest } |\nabla_{s_t}J(Q_t, Q_{t-1} + r_t)| \text{dimension}
\end{aligned}
\end{equation}

\vspace{1mm}\noindent\textbf{Attack with smaller frequency.} The strategically-timed attack~\citep{Lin2017Tacticsc} aims to decrease the attack frequency without sacrificing the performance of the un-targeted reinforcement attack.
We formally present it below:
\begin{equation}
\begin{aligned}
    & \delta_t = \delta_t * c_t \quad c_t \in \{0, 1\},\\
    & \frac{\sum_{t=1}^{T}c_t}{T} < d\\
\end{aligned}
\end{equation}
where $c_t$ is a binary variable that controls when to attack; $d < T$ is the frequency of adversarial examples. There are two approaches to generate the binary sequence $c_{1:T}$ optimizing a hard integer programming problem and generating sequences via heuristic methods. Let $p_0, p_1$ be the two maximum probability of an policy $\pi$, we define $c_t$ as follows, which is different from~\citep{Lin2017Tacticsc}:
$$
c_t = (p_0 - p_1) > threshold
$$
In our experiments, we let the RL-based recommendation system have a peak probability at the maximum action to test the importance of the action to attackers using the above formula.
In contrast to the above methods, Jacobian-based Saliency Map Attack (JSMA) and Deepfool are based on the gradient of actions rather than the gradient of $Q$ value. One key component of JSMA is saliency map computation used to decide which dimension of vectors (in Image classification is pixels) are modified. Deepfool pinpoints the attack dimension by comparison of affine distances between some class and temporal classes. More details can be found in \cite{papernot2016limitations} and \cite{moosavi2016deepfool}.

\vspace{1mm}\noindent\textbf{Counterfactual Based Attack.}
Counterfactual can find a similar version of the query input within some distributions, changing the model decisions and receiving a different classification. This helps to explain why a specific classification decision was made by a model and improve the interpretation of model boundaries~\citep{yang2021model}, which is known as counterfactual explanations. Recent work reveals that counterfactual explanations produce adversarial examples in deep neural networks~\citep{browne2020semantics}. Therefore, we propose to generate counterfactual data samples to be the counterfactual-based attack.

Most of the adversarial examples are generated by adding perturbations. The counterfactual-based attack is recognized as one sub-type of adversarial examples, which is different from traditional perturbations. One of the majority differences is that the counterfactual-based attack is generated by causal reasoning. To capture the casual relationships, we introduce Structural Causal Model(SCM) $M=\langle U,V,F\rangle$, given by a  directed acyclic graph (DAG) \( \mathcal{G} \), where:
\begin{itemize}
    \item $U=(U_1,...,U_N)$ is a set of exogenous variables determined by unobserved and omitted factors. We assume that these noises are independent variables such that $U_i$ is independent of all other noise variables.
    \item $V=(V_1,...,V_N)$ is a set of endogenous variables that are observed nodes in the DAG.
    \item $F=(f_1,...,f_N)$  is a set of structural equations representing the set of edges in the DAG. Each represents a causal mechanism that expresses the value of $V_i$ as a function of the values of parents of $V_i$ in \( \mathcal{G} \) and the noise term, that is $V_i = f_i(Pa(V_i)_i, U_i)$.
\end{itemize}
 

To simplify counterfactual reasoning, we assume that the input states follow the Local Causal Models (LCMs)~\citep{pitis2020counterfactual}. Hence, we assume the set of edges in \( \mathcal{G} \) satisfying the Structural Minimality hypothesis, stating that $V_j$ is a parent node of $V_i$ in \( \mathcal{G} \) if and only if there is a direct edge from $V_j$ to $V_i$ such that setting the value of $V_j$ will have a direct effect on $V_i$ through $f_i$. With this assumption, a large subspace \( \mathcal{L} \) often exists for each pair of nodes ($Pa(V_i)_j,V_i$) in the DAG, in which two components are causally independent conditioning on a subset of parents nodes of $V_i$ so that can be considered separately for training and inference.

Specifically, given two states with the same local factorization, we find the intersection of these two states. The intersection parts remain unchanged in the MDP process, representing the critical components containing user identifiable information. Leaving the critical components untouched, we produce a new counterfactual-based attack by swapping a subset of the other factors of two states under the assumption that two components are locally independent. The algorithm is given in Algorithm 1. This process can be interpreted by making intervention $do(S_t^{i,...,j}) = S_{t'}^{i',...,j'}$ on the Local Causal Models $M^{\mathcal{L}}$ to obtain the simulation result~\citep{pitis2020counterfactual}.

\begin{algorithm}[H]
    \SetKwInOut{Input}{input}
    \SetKwInOut{Output}{output}
    \SetAlgoLined
    \Input{Current state $s_i$ and a random state $s_j$}
    $s_{same} = s_i \cap s_j$\;
    $s_{ir} \leftarrow s_{i} \setminus s_{same}$ \;
    $s_{jr} \leftarrow s_{j} \setminus s_{same}$ \;
    $s_i \leftarrow s_{ir}\cup s_{same}$\;
    \Output{Counterfactual State $s_i$}
    
    \caption{Counterfactual State Generation}\label{alg:cap}
\end{algorithm}


\subsection{Detection Model} 
The detection model is a supervised classifier, which detects adversarial examples based on the actions of the reinforcement agent in a general feature space.
Suppose the action distributions of an agent are shifted by adversarial examples (Section \ref{exp} shows statistical evidence of the drift).
Given an abnormal action sequence $a = \pi^{*}(a|s + \delta)$ or a counterfactual action sequence, the detection model aims to establish a separating hyperplane between adversarial examples and normal examples, thereby measuring the probability $p(y | a, \theta)$ or $p(y | \pi^{*}, s, \delta, \theta)$, where $y$ is a binary variable indicating whether the input data are attacked.

To detect the adversarial examples presented in the last section, we employ an attention-based classifier. We first conduct statistical analysis on the attacked actions whose result is shown in section \ref{exp}. The detection model consists of two parts. The first is an encoder to encode the action methods into a low-dimensional feature vector. The second is a classifier to separate different data. We adopt this encoder-decoder model to make a bottleneck and filter out noisy information. The formulation of GRU is as follows:
\begin{equation}
\begin{aligned}
    & z_t = \sigma_g(W_z a_t + U_z h_{t-1})\\
    & r_t = \sigma_g(W_r a_t + U_r h_{t-1})\\
    & \hat{h_t} = tanh(W_h a_t + U_h \circ h_{t-1})\\
    & h_t = (1 - z_t) \circ h_{t-1} + z_t \circ \hat{h_t}
\end{aligned}
\end{equation}

We use an action sequence $a_{1:T}$ to denote a series of user relation vectors or item embedding vectors and apply a recurrent model to encode the temporal relation into the feature vectors.
We further adopt a single layer GRU network as our encoder and employ the attention-based dense net for detecting adversarial examples (formulated below).
\begin{equation}
\begin{aligned}
    & \alpha_{t} = Softmax(W_{e} e + b_{e})_{t}\\
    & att, hid = \sum\limits_{t=1}^{T} \alpha_{t} h_{t}\\
    & p = Softmax(W_{att} att + b_{att})
\end{aligned}
\end{equation}
where $e$ is the combined vector of action embedding and hidden states $hid$---we compute attention weights from embedding vectors and employ a liner unit to distribute probabilities to input time steps;
$h_{t}$ is the output of encoder. The vectors processed through the attention layer is then piped into a linear unit with softmax to compute the probability of adversarial examples. The loss function is the cross entropy between the true label and corresponding probability,
$$
J(Att(a_{1:T}), y) = - y \circ log(p)
$$

\section{Experiments}
\label{exp}
In this section, we report our experiments to evaluate attack methods and our detection model.
We first introduce the datasets and then provide quantitative evaluation and discussion on different attacks and our detection model.

\begin{figure*}[ht]
  \centering
    \includegraphics[width=\textwidth]{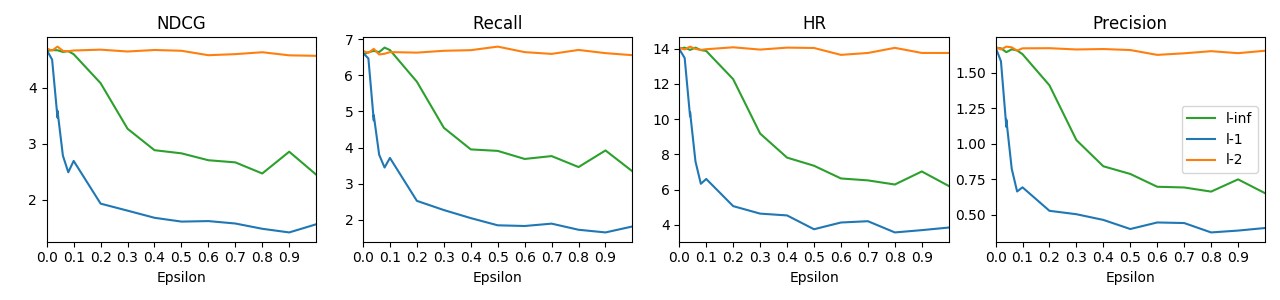}
    \includegraphics[width=\textwidth]{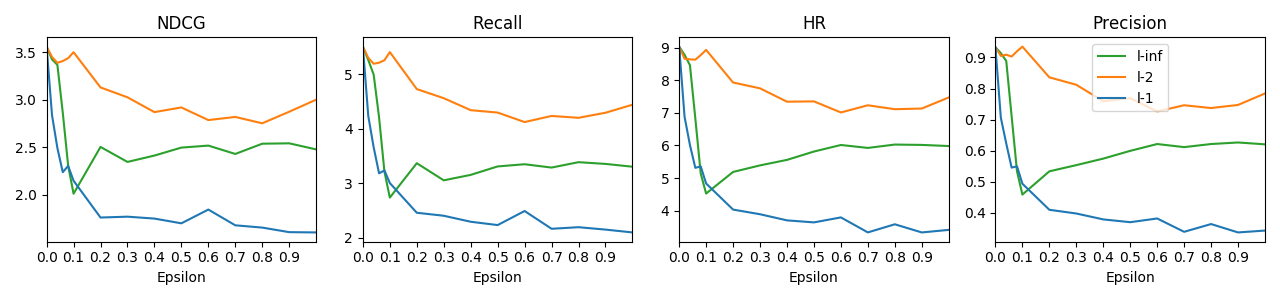}
    \includegraphics[width=\textwidth]{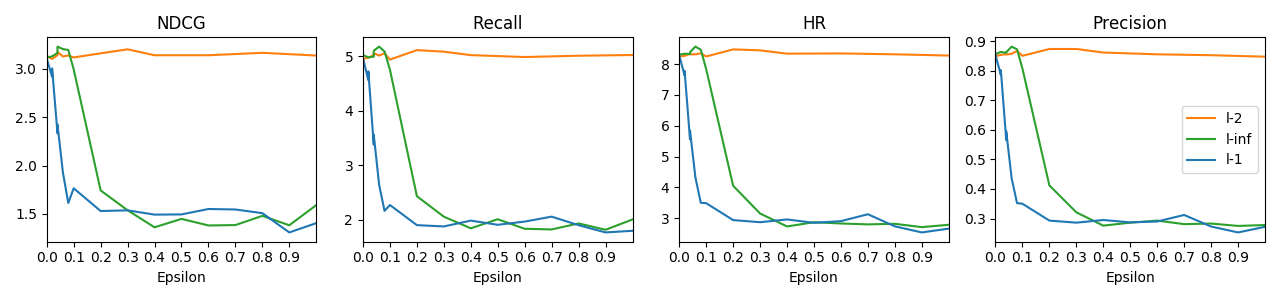}
  \centering
  \caption{Comparison of three attack methods, $l_{\infty}$, $l_2$, and $l_1$ on three datasets (from top to bottom): Amazon Beauty, Amazon Cellphones, and Amazon Clothing.
  }
  \label{fig-attack-0}
\end{figure*}

\subsection{Dataset and Experiment Setup}
\label{exp-dataset}
We conduct experiments based on two reinforcement learning interactive recommendation systems. Following~\cite{chen2019large} and~\cite{xian2019reinforcement} over the real-world dataset, Amazon dataset~\citep{he2016ups}. This public dataset contains user reviews and metadata of the Amazon e-commerce platform from 1996--2014. We utilize three subsets named Beauty, Cellphones, and Clothing as our dataset. We directly use the dataset provided by~\cite{xian2019reinforcement} on Github to reproduce their experiments. Details about Amazon dataset analysis can be found in~\cite{xian2019reinforcement}.


We conduct our attack and detection experiments based on~\cite{xian2019reinforcement}.
We preprocess the dataset by filtering out feature words with higher TF-IDF scores than 0.1.
Then, we use 70\% data in each dataset as the training set (and the rest as the test set) and actions of reinforcement agent as the detection data.
We define the actions of PGPR~\citep{xian2019reinforcement} as heterogeneous graph paths that start from users and have a length of 4.
The three Amazon sub-dataset (Beauty, Cellphones, and Clothing) contain 22,363, 27,879, and 39,387 users.
To accelerate experiments, the first 10,000 users of each dataset are extracted for adversarial example production. Users in Beauty get on average 127.51 paths. The counterparts for Cellphones and Clothing are 121.92 and 122.71. We adopt the action file of $l_{\infty}$ attack with an epsilon of 0.5 as the training set. As the number of paths is large, we utilize the first 100,000 paths for train and validation. The ratio of train validation is 80/20.  Regarding the test, 100,000 paths from each action file are randomly sampled as the test set. 


We slightly modify JSMA and Deepfool for our experiments---we create the saliency map by calculating the product of the target label and temporal label to achieve both effectiveness and higher efficiency (by 0.32 seconds per iteration) of JSMA;
we also use sampling to decrease the computation load on a group of gradients for Deepfool.
Besides, we set the hidden size of the GRU to 32 for the encoder, the drop rate of the attention-based classifier to 0.5, the maximum length of a user-item path to 4 (according to~\cite{xian2019reinforcement}), and the learning rate and weight decay of the optimization solver, Adam, to 5e-4 and 0.01, respectively.

\subsection{Attack Experiments}
This section reports our experiments on adversarial attacks. The first part shows the attack experiment results, followed by an analysis of the impact of attack frequency, attack intensity, and the action space of the recommendation system on the attack performance.

\vspace{1mm}\noindent\textbf{Adversarial attack results.} We are interested in how vulnerable the agent is to perturbation in semantic embedding space. We consider an attack to be effective if a small perturbation leads to a notable performance reduction. We experimentally compare the performance of different attack methods (described in Section \ref{method}) in Table \ref{tab-basic-attack}. 

We reuse the evaluation metrics of the original model, namely Normalized Discounted Cumulative Gain (NDCG), Recall, Hit Ratio (HR), and Precision for evaluation on the amazon dataset.
Table \ref{tab-basic-attack} shows the attack results share the same trend with the distribution discrepancy in Table \ref{tab-mmd}. Most attack methods significantly reduce the performance of the reinforcement system. FGSM $l_1$ achieves the best performance. It reveals that attacks on a single dimension can change the neural network's action drastically. Compared with $l_1$ and $l_{inf}$ methods, FGSM $l_2$ is less effective on three datasets, where the evaluation metrics are mostly same in contrast to the case without attack (The original baseline in Table \ref{tab-mmd}). 
It is worth mentioning that counterfactual attack does not perform well as the others. One of the possible reasons is that the generated counterfactual state still falls in the original latent space. The counterfactual attack can introduce noise to the current state by introducing irrelevant information from future states.

\begin{table}[ht]
  \caption{Adversarial attack results on Amazon Beauty}
  \label{tab-basic-attack}
  \begin{tabular}{lrrrrr}
    \hline
    Data &Parameters &NDCG &Recall &HR &Precision\\\hline

    Original &- &5.449 &8.324 &14.401 &1.707\\
    FGSM $l_1$ &$\epsilon=$0.1 &2.695 &3.714 &6.599 &0.693\\
    FGSM $l_2$ &$\epsilon=$1.0 &4.567 &6.555 &13.751 &1.653\\
    FGSM $l_{inf}$ &$\epsilon=$0.5 &2.830 &3.909 &7.351 &0.787 \\
    Counterfactual &- &5.324 &8.089 &14.077 &1.658  \\
    JSMA &- &2.984 &3.844 &8.254 &0.931\\
    Deepfool &- &3.280 &4.352 &9.548 &1.050\\\hline
\end{tabular}
\end{table}

 
\vspace{1mm}\noindent\textbf{Impact of attack intensity.} Adversarial examples make small perturbations to achieve notable changes of recommendation performance. Although larger perturbations on user-item interaction embeddings are not easily perceptible by humans, decreasing attack intensity might degrade attack effectiveness. To demonstrate the impact of different attack intensities in the context of RL-based recommender systems, we conduct the empirical experiment by varying the attack intensity, which is reflected by the $\epsilon$ parameter shown in Equation \ref{eqt-l-inf} and Equation \ref{eqt-l1}.
Experiment results of attack with epsilon variation of FGSM attack methods on three Amazon datasets (Figure \ref{fig-attack-0}) show that compared to a $0.0$ value epsilon, all metric values decline as Epsilon increases, and $l_1$ attack achieves the best result.
$l_1$ follows a similar yet more abrupt trend than the $l_{\infty}$ attack, while
the $l_2$ attack achieves the worst performance regardless of the epsilon value.
\cite{Huang2017Adversarial} propose to attack the reinforcement learning applied on games such as Atari. Their experiments reveal that the $l_2$ attack achieves comparable performance as $l_1$ and $l_{inf}$ attack do. To exclude the possibility that the $l_2$ might be more effective with larger epsilon values, we set $\epsilon$ to 20.0 to test if $l_2$, but the result is the same. This observation that the attack in user-item-feature embedding space shows different characteristics from attacks in the pixel space.

Another interesting observation is that the metric values show different trends depending on the datasets---unlike on Beauty and Cellphones, the $l_{\infty}$ attack achieves comparable performance to $l_1$ on the Clothing dataset when the $\epsilon$ is larger than 0.3. The result on the Cellphones dataset shows that the effectiveness of the $l_{inf}$ attack diminishes as the $\epsilon$ continues increasing beyond 0.1.

\begin{figure}[!htb]
  \centering
    \includegraphics[width=\textwidth]{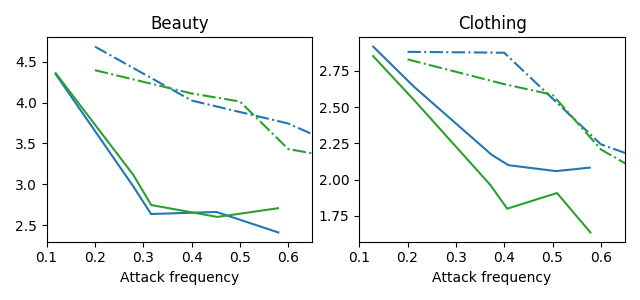}
  \centering
  \caption{NDCG of attack frequency on Amazon Beauty and Clothing. Dashdot lines represent random attacks, solid lines are strategically-timed attacks. Blue lines are FGSM $l_{inf}$ attacks, green lines are FGSM $l_1$ attacks.}
  \label{fig-attack-time}
\end{figure}

\vspace{1mm}\noindent\textbf{Impact of attack frequency.}
We conduct two experiments on attack frequency, random attack and strategic attack. In the random attack method, the adversarial examples are crafted with a frequency parameter, $p_{freq}$. In the strategically-timed attack, the adversarial examples are generated by the method shown in Section \ref{method-attack}. The NDCG metric is presented in Figure \ref{fig-attack-time}; other metrics have a similar trend. It can be seen from \ref{fig-attack-time} that the random attack performs worse than the strategically-timed attack. Generating strategically adversarial examples one third to half time steps achieves a significant reduction in all metrics.

\subsection{Detection Experiments}
\vspace{1mm}\noindent\textbf{Analysis of adversarial examples.}
We use Maximum Mean Discrepancy as statistical measures of high dimensional data distribution distance. This divergence is defined as:
$$
MMD(k, X_{org}, X_{adv}) = \sup\limits_{k \in K}\Bigg(\frac{1}{n}\sum\limits_{i=1}^{n}k(x_{org,i}) - \frac{1}{m}\sum\limits_{i=1}^{n}k(x_{adv,i})\Bigg)
$$
where $k$ is the kernel function, i.e., a radial basis function, which measures the distance between the means of two distributions (Table \ref{tab-mmd} shows the results); $X_{org}, X_{adv}$ are benign and adversarial examples. The data is randomly sampled from generated action embeddings of interactive recommendation systems. Each MMD is computed by averaging 40 batches of 500 samples. The actions generated by the RL agent are paths of the users-relation-item graph. As mentioned in Section \ref{exp-dataset}, each user gets over 100 paths, which determines the overlapping of original data and adversarial examples decrease with the time step of the path growing. We choose the embedding of the last step to represent the recommended items.

MMD-org shows the discrepancy between the original and adversarial datasets,  where MMD-$l_1$ presents the discrepancy between different attack methods. The results (Table \ref{tab-mmd}) show that the adversarial distribution is different from the original distribution. Also, the disturbed distributions are closed to each other regardless of the attack type.
This insight clarifies that we can use a classifier to separate benign data and adversarial data, and it can detect several attacks simultaneously, which might be transferred to other reinforcement learning attack detection tasks.

\begin{table}
  \caption{MMD between benign distribution and adversarial distribution on Amazon Beauty}
  \label{tab-mmd}
  \begin{tabular}{lrrr}
    \hline 
    Data &Parameters &MMD-org &MMD-$l_1$\\ \hline 
    Original &- &0.121 &0.620\\
    FGSM $l_1$ &$\epsilon=$0.1 & 0.604 &0.010\\
    FGSM $l_2$ &$\epsilon=$1.0 & 0.016 &0.573\\
    FGSM $l_{inf}$ &$\epsilon=$0.5 & 0.570 &0.011\\
    Counterfactual &- & 0.232 &0.273\\
    JSMA &- &0.412 &0.034\\
    Deepfool &- &0.177 &0.458\\ \hline
\end{tabular}
\end{table}


\vspace{1mm}\noindent\textbf{Detection Performance.} 
From a statistical perspective, above analysis shows that one classifier can detect multiple types of attacks.
We evaluate the detection performance of different models using Precision, Recall, and F1 score.

We adopt an attention-based network for detection experiments. The detection model is trained on FGSM $l_1$ attack with $\epsilon$ at 0.1 for all datasets. The results (Table \ref{tab-detection-basic}) show that our detection model achieves better performance on attacks that cause serious disruption. The detection precision and recall rise as the attack is stronger.
$l_{\infty}$ attack validates this trend, which shows that our model can detect weaker attacks as well. The result of detection on $l_2$ attack can be reasoned with the MMD analysis shown above. High precision and low recall show that most $l_2$ adversarial examples are close to benign data, which confuses the detector.
The $l_1$ attack with $\epsilon=1.0$ validates our detector performs well yet achieves worse performance on other tests on the Cellphones dataset.
Our model can also detect the counterfactual-based attack since the data distribution has been changed, verifying that our detection model can detect different types of attacks.
Our results on factor analysis (Table \ref{tab-detection-basic}) show that the detection model can detect attacks even under low attack frequencies.
But the detection accuracy decreases as the attack frequency drops---the recall decreases significantly to 40.1\% when 11.8\% examples represent attacks.

\begin{table}[!htb]
  \caption{Detection Result and Factor Analysis}
  \label{tab-detection-basic}
  \begin{tabular}{lrrrr}
    \hline
    Dataset &Attack &Precision &Recall &F1 Score\\
    \hline
    Beauty &$l_1$ 0.1 &0.919 &0.890 &0.904\\
           &$l_2$ 1.0 &0.605 &0.119 &0.199\\
           &$l_{inf}$ 0.5 &0.918 &0.871 &0.894\\
           &Counterfactual &0.900 &0.895 &0.898\\
           &JSMA &0.910 &0.793 &0.848\\
           &Deepfool &0.915 &0.840 &0.876\\
    \hline
    Cellphones &$l_1$ 0.1 &0.801 &0.781 &0.791\\
           &$l_2$ 1.0 &0.754 &0.593 &0.664\\
           &$l_{inf}$ 0.5 &0.795 &0.752 &0.773\\
           &$l_1$ 1.0 &0.810 &0.825 &0.817\\
    \hline
    Clothing &$l_1$ 0.1 &0.911 &0.866 &0.888\\
           &$l_2$ 1.0 &0.541 &0.099 &0.168\\
           &$l_{inf}$ 0.5 &0.912 &0.879 &0.895\\
    \hline
    Dataset &Frequency &Precision &Recall &F1 Score\\
    \hline
    Beauty &$l_1$ 0.02 &0.823 &0.362 &0.503\\
           &$l_1$ 0.04 &0.906 &0.754 &0.823\\
           &$l_1$ 0.06 &0.915 &0.841 &0.876\\
           &$l_1$ 0.08 &0.918 &0.872 &0.894\\
           &$l_1$ 0.3 &0.922 &0.927 &0.924\\
    \hline
    Dataset &Frequency &Precision &Recall &F1 Score\\
    \hline
    Beauty &$l_1$ 0.579 &0.921 &0.912 &0.917\\
           &$l_1$ 0.451 &0.921 &0.908 &0.914\\
           &$l_1$ 0.316 &0.918 &0.879 &0.898\\
           &$l_1$ 0.279 &0.914 &0.829 &0.869\\
           &$l_1$ 0.118 &0.837 &0.401 &0.543\\
  \hline 
\end{tabular}
\end{table}

\section{Conclusion}
Adversarial attacks on reinforcement agents can greatly degrade user experience in interactive recommendation systems, as an intervention on causal factors can result in a different recommended result. In this paper, we systematically study adversarial attacks on reinforcement learning-based recommendation systems by investigating different attack methods and the impact of attack intensity and frequency on the performance of adversarial examples. We conduct statistical analysis to show that classifiers, especially an attention-based detector, can well separate the detection data. Our extensive experiments show the excellent performance of our attack and detection models.

\bibliographystyle{frontiersinSCNS_ENG_HUMS} 
\bibliography{frontiers}

\begin{thebibliography}{32}
\providecommand{\natexlab}[1]{#1}
\expandafter\ifx\csname urlstyle\endcsname\relax
  \providecommand{\doi}[1]{doi:\discretionary{}{}{}#1}\else
  \providecommand{\doi}{doi:\discretionary{}{}{}\begingroup
  \urlstyle{rm}\Url}\fi
\providecommand{\selectlanguage}[1]{\relax}
\providecommand{\bibAnnoteFile}[1]{%
  \IfFileExists{#1}{\begin{quotation}\noindent\textsc{Key:} #1\\
  \textsc{Annotation:}\ \input{#1}\end{quotation}}{}}
\providecommand{\bibAnnote}[2]{%
  \begin{quotation}\noindent\textsc{Key:} #1\\
  \textsc{Annotation:}\ #2\end{quotation}}

\bibitem[{Bendale and Boult(2016)}]{Bendale2016Open}
Bendale, A. and Boult, T.~E. (2016).
\newblock Towards {{Open Set Deep Networks}}.
\newblock In \emph{2016 {{IEEE Conference}} on {{Computer Vision}} and
  {{Pattern Recognition}} ({{CVPR}})} ({IEEE}), 1563--1572
\bibAnnoteFile{Bendale2016Open}

\bibitem[{Browne and Swift(2020)}]{browne2020semantics}
Browne, K. and Swift, B. (2020).
\newblock Semantics and explanation: why counterfactual explanations produce
  adversarial examples in deep neural networks.
\newblock \emph{arXiv preprint arXiv:2012.10076}
\bibAnnoteFile{browne2020semantics}

\bibitem[{Carlini et~al.(2019)Carlini, Athalye, Papernot, Brendel, Rauber,
  Tsipras et~al.}]{carlini2019Onevaluating}
Carlini, N., Athalye, A., Papernot, N., Brendel, W., Rauber, J., Tsipras, D.,
  et~al. (2019).
\newblock On evaluating adversarial robustness.
\newblock \emph{arXiv preprint arXiv:1902.06705}
\bibAnnoteFile{carlini2019Onevaluating}

\bibitem[{Carlini and Wagner(2017)}]{Carlini2017Adversariala}
Carlini, N. and Wagner, D. (2017).
\newblock Adversarial examples are not easily detected: Bypassing ten detection
  methods.
\newblock In \emph{Proceedings of the 10th ACM Workshop on Artificial
  Intelligence and Security} (ACM), 3--14
\bibAnnoteFile{Carlini2017Adversariala}

\bibitem[{Chen et~al.(2019{\natexlab{a}})Chen, Dai, Cai, Zhang, Wang, Tang
  et~al.}]{chen2019large}
Chen, H., Dai, X., Cai, H., Zhang, W., Wang, X., Tang, R., et~al.
  (2019{\natexlab{a}}).
\newblock Large-scale interactive recommendation with tree-structured policy
  gradient.
\newblock In \emph{Proceedings of the AAAI Conference on Artificial
  Intelligence} (AAAI), vol.~33, 3312--3320
\bibAnnoteFile{chen2019large}

\bibitem[{Chen et~al.(2019{\natexlab{b}})Chen, Carlini, and
  Wagner}]{Chen2019Stateful}
Chen, S., Carlini, N., and Wagner, D. (2019{\natexlab{b}}).
\newblock Stateful {{Detection}} of {{Black}}-{{Box Adversarial Attacks}}
\bibAnnoteFile{Chen2019Stateful}

\bibitem[{Christakopoulou et~al.(2018)Christakopoulou, Beutel, Li, Jain, and
  Chi}]{christakopoulou2018q}
Christakopoulou, K., Beutel, A., Li, R., Jain, S., and Chi, E.~H. (2018).
\newblock Q\&r: A two-stage approach toward interactive recommendation.
\newblock In \emph{Proceedings of the 24th ACM SIGKDD International Conference
  on Knowledge Discovery \& Data Mining} (ACM), 139--148
\bibAnnoteFile{christakopoulou2018q}

\bibitem[{G.Adomavicius and A.Tuzhilin(2005)}]{adomavicius2005toward}
G.Adomavicius and A.Tuzhilin (2005).
\newblock Toward the next generation of recommender systems: a survey of the
  state-of-the-art and possible extensions.
\newblock \emph{IEEE Transactions on Knowledge and Data Engineering} 17,
  734--749.
\newblock \doi{10.1109/TKDE.2005.99}
\bibAnnoteFile{adomavicius2005toward}

\bibitem[{Gao et~al.(2018)Gao, Lanchantin, Soffa, and Qi}]{gao2018black}
Gao, J., Lanchantin, J., Soffa, M.~L., and Qi, Y. (2018).
\newblock Black-box generation of adversarial text sequences to evade deep
  learning classifiers.
\newblock In \emph{2018 IEEE Security and Privacy Workshops (SPW)} (IEEE),
  50--56
\bibAnnoteFile{gao2018black}

\bibitem[{Gleave et~al.(2019)Gleave, Dennis, Kant, Wild, Levine, and
  Russell}]{Gleave2019Adversariala}
Gleave, A., Dennis, M., Kant, N., Wild, C., Levine, S., and Russell, S. (2019).
\newblock Adversarial {{Policies}}: {{Attacking Deep Reinforcement Learning}}
\bibAnnoteFile{Gleave2019Adversariala}

\bibitem[{Goodfellow et~al.(2014)Goodfellow, Shlens, and
  Szegedy}]{Goodfellow2014Explaininga}
Goodfellow, I.~J., Shlens, J., and Szegedy, C. (2014).
\newblock Explaining and {{Harnessing Adversarial Examples}}
\bibAnnoteFile{Goodfellow2014Explaininga}

\bibitem[{Grosse et~al.(2017)Grosse, Manoharan, Papernot, Backes, and
  McDaniel}]{Grosse2017(Statistical)}
Grosse, K., Manoharan, P., Papernot, N., Backes, M., and McDaniel, P. (2017).
\newblock On the ({{Statistical}}) {{Detection}} of {{Adversarial Examples}}
\bibAnnoteFile{Grosse2017(Statistical)}

\bibitem[{He and McAuley(2016)}]{he2016ups}
He, R. and McAuley, J. (2016).
\newblock Ups and downs: Modeling the visual evolution of fashion trends with
  one-class collaborative filtering.
\newblock In \emph{proceedings of the 25th international conference on world
  wide web} (International World Wide Web Conferences Steering Committee),
  507--517
\bibAnnoteFile{he2016ups}

\bibitem[{Hendrycks and Gimpel(2016)}]{Hendrycks2016Early}
Hendrycks, D. and Gimpel, K. (2016).
\newblock Early {{Methods}} for {{Detecting Adversarial Images}}
\bibAnnoteFile{Hendrycks2016Early}

\bibitem[{Huang et~al.(2017)Huang, Papernot, Goodfellow, Duan, and
  Abbeel}]{Huang2017Adversarial}
Huang, S., Papernot, N., Goodfellow, I., Duan, Y., and Abbeel, P. (2017).
\newblock Adversarial {{Attacks}} on {{Neural Network Policies}}
\bibAnnoteFile{Huang2017Adversarial}

\bibitem[{Kos and Song(2017)}]{Kos2017Delving}
Kos, J. and Song, D. (2017).
\newblock Delving into adversarial attacks on deep policies
\bibAnnoteFile{Kos2017Delving}

\bibitem[{Lin et~al.(2017)Lin, Hong, Liao, Shih, Liu, and
  Sun}]{Lin2017Tacticsc}
Lin, Y.-C., Hong, Z.-W., Liao, Y.-H., Shih, M.-L., Liu, M.-Y., and Sun, M.
  (2017).
\newblock Tactics of {{Adversarial Attack}} on {{Deep Reinforcement Learning
  Agents}}
\bibAnnoteFile{Lin2017Tacticsc}

\bibitem[{Mahmood and Ricci(2007)}]{mahmood2007learning}
Mahmood, T. and Ricci, F. (2007).
\newblock Learning and adaptivity in interactive recommender systems.
\newblock In \emph{Proceedings of the ninth international conference on
  Electronic commerce} (ACM), 75--84
\bibAnnoteFile{mahmood2007learning}

\bibitem[{Mahmood and Ricci(2009)}]{Mahmood2009Improving}
Mahmood, T. and Ricci, F. (2009).
\newblock Improving recommender systems with adaptive conversational
  strategies.
\newblock In \emph{Proceedings of the 20th {{ACM}} Conference on {{Hypertext}}
  and Hypermedia - {{HT}} '09} ({ACM Press}), 73
\bibAnnoteFile{Mahmood2009Improving}

\bibitem[{Metzen et~al.(2017)Metzen, Genewein, Fischer, and
  Bischoff}]{Metzen2017Detecting}
Metzen, J.~H., Genewein, T., Fischer, V., and Bischoff, B. (2017).
\newblock On {{Detecting Adversarial Perturbations}}
\bibAnnoteFile{Metzen2017Detecting}

\bibitem[{Moosavi-Dezfooli et~al.(2016)Moosavi-Dezfooli, Fawzi, and
  Frossard}]{moosavi2016deepfool}
Moosavi-Dezfooli, S.-M., Fawzi, A., and Frossard, P. (2016).
\newblock Deepfool: a simple and accurate method to fool deep neural networks.
\newblock In \emph{Proceedings of the IEEE conference on computer vision and
  pattern recognition}. 2574--2582
\bibAnnoteFile{moosavi2016deepfool}

\bibitem[{Papernot et~al.(2016)Papernot, McDaniel, Jha, Fredrikson, Celik, and
  Swami}]{papernot2016limitations}
Papernot, N., McDaniel, P., Jha, S., Fredrikson, M., Celik, Z.~B., and Swami,
  A. (2016).
\newblock The limitations of deep learning in adversarial settings.
\newblock In \emph{2016 IEEE European Symposium on Security and Privacy
  (EuroS\&P)} (IEEE), 372--387
\bibAnnoteFile{papernot2016limitations}

\bibitem[{Pattanaik et~al.(2017)Pattanaik, Tang, Liu, Bommannan, and
  Chowdhary}]{Pattanaik2017Robust}
Pattanaik, A., Tang, Z., Liu, S., Bommannan, G., and Chowdhary, G. (2017).
\newblock Robust {{Deep Reinforcement Learning}} with {{Adversarial Attacks}}
\bibAnnoteFile{Pattanaik2017Robust}

\bibitem[{Pinto et~al.(2017)Pinto, Davidson, Sukthankar, and
  Gupta}]{Pinto2017Robusta}
Pinto, L., Davidson, J., Sukthankar, R., and Gupta, A. (2017).
\newblock Robust {{Adversarial Reinforcement Learning}}
\bibAnnoteFile{Pinto2017Robusta}

\bibitem[{Pitis et~al.(2020)Pitis, Creager, and Garg}]{pitis2020counterfactual}
Pitis, S., Creager, E., and Garg, A. (2020).
\newblock Counterfactual data augmentation using locally factored dynamics.
\newblock \emph{arXiv preprint arXiv:2007.02863}
\bibAnnoteFile{pitis2020counterfactual}

\bibitem[{Szegedy et~al.(2013)Szegedy, Zaremba, Sutskever, Bruna, Erhan,
  Goodfellow et~al.}]{Szegedy2013Intriguing}
Szegedy, C., Zaremba, W., Sutskever, I., Bruna, J., Erhan, D., Goodfellow, I.,
  et~al. (2013).
\newblock Intriguing properties of neural networks
\bibAnnoteFile{Szegedy2013Intriguing}

\bibitem[{Taghipour and Kardan(2008)}]{Taghipour2008hybrid}
Taghipour, N. and Kardan, A. (2008).
\newblock A hybrid web recommender system based on q-learning.
\newblock In \emph{Proceedings of the 2008 ACM symposium on Applied computing}
  (ACM), 1164--1168
\bibAnnoteFile{Taghipour2008hybrid}

\bibitem[{Taghipour et~al.(2007)Taghipour, Kardan, and
  Ghidary}]{Taghipour2007Usage-Based}
Taghipour, N., Kardan, A., and Ghidary, S.~S. (2007).
\newblock Usage-based web recommendations: a reinforcement learning approach.
\newblock In \emph{Proceedings of the 2007 ACM conference on Recommender
  systems} (ACM), 113--120
\bibAnnoteFile{Taghipour2007Usage-Based}

\bibitem[{Thompson et~al.(2004)Thompson, Goker, and
  Langley}]{thompson2004personalized}
Thompson, C.~A., Goker, M.~H., and Langley, P. (2004).
\newblock A personalized system for conversational recommendations.
\newblock \emph{Journal of Artificial Intelligence Research} 21, 393--428
\bibAnnoteFile{thompson2004personalized}

\bibitem[{Xian et~al.(2019)Xian, Fu, Muthukrishnan, de~Melo, and
  Zhang}]{xian2019reinforcement}
Xian, Y., Fu, Z., Muthukrishnan, S., de~Melo, G., and Zhang, Y. (2019).
\newblock Reinforcement knowledge graph reasoning for explainable
  recommendation.
\newblock \emph{arXiv preprint arXiv:1906.05237}
\bibAnnoteFile{xian2019reinforcement}

\bibitem[{Yang et~al.(2021)Yang, Alva, Chen, and Hu}]{yang2021model}
Yang, F., Alva, S.~S., Chen, J., and Hu, X. (2021).
\newblock Model-based counterfactual synthesizer for interpretation.
\newblock \emph{arXiv preprint arXiv:2106.08971}
\bibAnnoteFile{yang2021model}

\bibitem[{Zhang et~al.(2019)Zhang, Yao, Sun, and Tay}]{zhang2019deep}
Zhang, S., Yao, L., Sun, A., and Tay, Y. (2019).
\newblock Deep learning based recommender system: A survey and new
  perspectives.
\newblock \emph{ACM Computing Surveys (CSUR)} 52, 5
\bibAnnoteFile{zhang2019deep}

\end{thebibliography}



%
%

\end{document}